# Artificial Neural Networks for Beginners

Carlos Gershenson
C.Gershenson@sussex.ac.uk

## 1. Introduction

The scope of this teaching package is to make a brief induction to Artificial Neural Networks (ANNs) for people who have no previous knowledge of them. We first make a brief introduction to models of networks, for then describing in general terms ANNs. As an application, we explain the backpropagation algorithm, since it is widely used and many other algorithms are derived from it.

The user should know algebra and the handling of functions and vectors. Differential calculus is recommendable, but not necessary. The contents of this package should be understood by people with high school education. It would be useful for people who are just curious about what are ANNs, or for people who want to become familiar with them, so when they study them more fully, they will already have clear notions of ANNs. Also, people who only want to apply the backpropagation algorithm without a detailed and formal explanation of it will find this material useful. This work should not be seen as "Nets for dummies", but of course it is not a treatise. Much of the formality is skipped for the sake of simplicity. Detailed explanations and demonstrations can be found in the referred readings. The included exercises complement the understanding of the theory. The on-line resources are highly recommended for extending this brief induction.

## 2. Networks

One efficient way of solving complex problems is following the lemma "divide and conquer". A complex system may be decomposed into simpler elements, in order to be able to understand it. Also simple elements may be gathered to produce a complex system (Bar Yam, 1997). Networks are one approach for achieving this. There are a large number of different types of networks, but they all are characterized by the following components: a set of nodes, and connections between nodes.

The nodes can be seen as computational units. They receive inputs, and process them to obtain an output. This processing might be very simple (such as summing the inputs), or quite complex (a node might contain another network...)

The connections determine the information flow between nodes. They can be unidirectional, when the information flows only in one sense, and bidirectional, when the information flows in either sense.

The interactions of nodes though the connections lead to a global behaviour of the network, which cannot be observed in the elements of the network. This global behaviour is said to be *emergent*. This means that the abilities of the network supercede the ones of its elements, making networks a very powerful tool.

Networks are used to model a wide range of phenomena in physics, computer science, biochemistry, ethology, mathematics, sociology, economics, telecommunications, and many other areas. This is because many systems can be seen as a network: proteins, computers, communities, etc. Which other systems could you see as a network? Why?

## 3. Artificial neural networks

One type of network sees the nodes as 'artificial neurons'. These are called artificial neural networks (ANNs). An artificial neuron is a computational model inspired in the natural neurons. Natural neurons receive signals through *synapses* located on the dendrites or membrane of the neuron. When the signals received are strong enough (surpass a certain *threshold*), the neuron is *activated* and emits a signal though the *axon*. This signal might be sent to another synapse, and might activate other neurons.

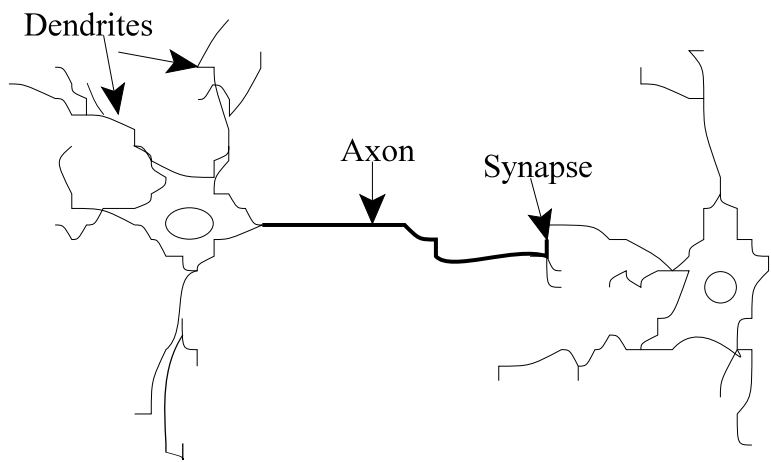

Figure 1. Natural neurons (artist's conception).

The complexity of real neurons is highly abstracted when modelling artificial neurons. These basically consist of *inputs* (like synapses), which are multiplied by *weights* (strength of the respective signals), and then computed by a mathematical function which determines the *activation* of the neuron. Another function (which may be the identity) computes the *output* of the artificial neuron (sometimes in dependance of a certain *threshold*). ANNs combine artificial neurons in order to process information.

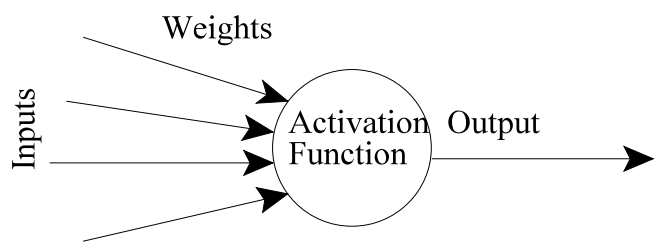

Figure 2. An artificial neuron

The higher a weight of an artificial neuron is, the stronger the input which is multiplied by it will be. Weights can also be negative, so we can say that the signal is *inhibited* by the negative weight. Depending on the weights, the computation of the neuron will be different. By adjusting the weights of an artificial neuron we can obtain the output we want for specific inputs. But when we have an ANN of hundreds or thousands of neurons, it would be quite complicated to find by hand all the necessary weights. But we can find algorithms which can adjust the weights of the ANN in order to obtain the desired output from the network. This process of adjusting the weights is called *learning* or *training*.

The number of types of ANNs and their uses is very high. Since the first neural model by McCulloch and Pitts (1943) there have been developed hundreds of different models considered as ANNs. The differences in them might be the functions, the accepted values, the topology, the learning algorithms, etc. Also there are many hybrid models where each neuron has more properties than the ones we are reviewing here. Because of matters of space, we will present only an ANN which learns using the backpropagation algorithm (Rumelhart and McClelland, 1986) for learning the appropriate weights, since it is one of the most common models used in ANNs, and many others are based on it.

Since the function of ANNs is to process information, they are used mainly in fields related with it. There are a wide variety of ANNs that are used to model real neural networks, and study behaviour and control in animals and machines, but also there are ANNs which are used for engineering purposes, such as pattern recognition, forecasting, and data compression.

**3.1. Exercise**

This exercise is to become familiar with artificial neural network concepts. Build a network consisting of four artificial neurons. Two neurons receive inputs to the network, and the other two give outputs from the network.

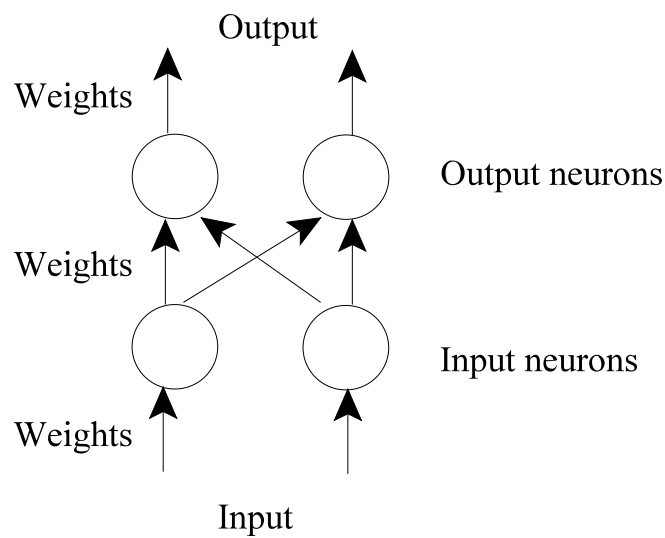

There are weights assigned with each arrow, which represent information flow. These weights are multiplied by the values which go through each arrow, to give more or less strength to the signal which they transmit. The neurons of this network just sum their inputs. Since the input neurons have only one input, their output will be the input they received multiplied by a weight. What happens if this weight is negative? What happens if this weight is zero?

The neurons on the output layer receive the outputs of both input neurons, multiplied by their respective weights, and sum them. They give an output which is multiplied by another weight.

Now, set all the weights to be equal to one. This means that the information will flow unaffected. Compute the outputs of the network for the following inputs: (1,1), (1,0), (0,1), (0,0), (-1,1), (-1,-1).

Good. Now, choose weights among 0.5, 0, and -0.5, and set them randomly along the network. Compute the outputs for the same inputs as above. Change some weights and see how the behaviour of the networks changes. Which weights are more critical (if you change those weights, the outputs will change more dramatically)?

Now, suppose we want a network like the one we are working with, such that the outputs should be the inputs in inverse order (*e.g.* (0.3,0.7)->(0.7,0.3)).

That was an easy one! Another easy network would be one where the outputs should be the double of the inputs.

Now, let's set thresholds to the neurons. This is, if the previous output of the neuron (weighted sum of the inputs) is greater than the threshold of the neuron, the output of the neuron will be one, and zero otherwise. Set thresholds to a couple of the already developed networks, and see how this affects their behaviour.

Now, suppose we have a network which will receive for inputs only zeroes and/or ones. Adjust the weights and thresholds of the neurons so that the output of the first output neuron will be the conjunction (AND) of the network inputs (one when both inputs are one, zero otherwise), and the output of the second output neuron will be the disjunction (OR) of the network inputs (zero in both inputs are zeroes, one otherwise). You can see that there is more than one network which will give the requested result.

Now, perhaps it is not so complicated to adjust the weights of such a small network, but also the capabilities of this are quite limited. If we need a network of hundreds of neurons, how would you adjust the weights to obtain the desired output? There are methods for finding them, and now we will expose the most common one.

## 4. The Backpropagation Algorithm

The backpropagation algorithm (Rumelhart and McClelland, 1986) is used in layered feed-forward ANNs. This means that the artificial neurons are organized in layers, and send their signals "forward", and then the errors are propagated backwards. The network receives inputs by neurons in the *input layer*, and the output of the network is given by the neurons on an *output layer*. There may be one or more intermediate *hidden layers*. The backpropagation algorithm uses supervised learning, which means that we provide the algorithm with examples of the inputs and outputs we want the network to compute, and then the error (difference between actual and expected results) is calculated. The idea of the backpropagation algorithm is to reduce this error, until the ANN *learns* the training data. The training begins with random weights, and the goal is to adjust them so that the error will be minimal.

The activation function of the artificial neurons in ANNs implementing the backpropagation algorithm is a weighted sum (the sum of the inputs $x_i$ multiplied by their respective weights $w_{ji}$):

$$A_j(\bar{x}, \bar{w}) = \sum_{i=0}^{n} x_i w_{ji} \qquad (1)$$

We can see that the activation depends only on the inputs and the weights.

If the output function would be the identity (output=activation), then the neuron would be called linear. But these have severe limitations. The most common output function is the sigmoidal function:

$$O_j(\bar{x}, \bar{w}) = \frac{1}{1 + e^{A_j(\bar{x}, \bar{w})}} \qquad (2)$$

The sigmoidal function is very close to one for large positive numbers, 0.5 at zero, and very close to zero for large negative numbers. This allows a smooth transition between the low and high output of the neuron (close to zero or close to one). We can see that the output depends only in the activation, which in turn depends on the values of the inputs and their respective weights.

Now, the goal of the training process is to obtain a desired output when certain inputs are given. Since the error is the difference between the actual and the desired output, the error depends on the weights, and we need to adjust the weights in order to minimize the error. We can define the error function for the output of each neuron:

$$E_j(\bar{x}, \bar{w}, d) = \left(O_j(\bar{x}, \bar{w}) - d_j\right)^2 \qquad (3)$$

We take the square of the difference between the output and the desired target because it will be always positive, and because it will be greater if the difference is big, and lesser if the difference is small. The error of the network will simply be the sum of the errors of all the neurons in the output layer:

$$E(\bar{x}, \bar{w}, \bar{d}) = \sum_j \left(O_j(\bar{x}, \bar{w}) - d_j\right)^2 \qquad (4)$$

The backpropagation algorithm now calculates how the error depends on the output, inputs, and weights. After we find this, we can adjust the weights using the method of *gradient descendent*:

$$\Delta w_{ji} = -\eta \frac{\partial E}{\partial w_{ji}} \qquad (5)$$

This formula can be interpreted in the following way: the adjustment of each weight ($\Delta w_{ji}$) will be the negative of a constant eta ($\eta$) multiplied by the dependance of the previous weight on the error of the network, which is the derivative of E in respect to $w_i$. The size of the adjustment will depend on $\eta$, and on the contribution of the weight to the error of the function. This is, if the weight contributes a lot to the error, the adjustment will be greater than if it contributes in a smaller amount. (5) is used until we find appropriate weights (the error is minimal). If you do not know derivatives, don't worry, you can see them now as functions that we will replace right away with algebraic expressions. If you understand derivatives, derive the expressions yourself and compare your results with the ones presented here. If you are searching for a mathematical proof of the backpropagation algorithm, you are advised to check it in the suggested reading, since this is out of the scope of this material.

So, we "only" need to find the derivative of E in respect to $w_{ji}$. This is the goal of the backpropagation algorithm, since we need to achieve this backwards. First, we need to calculate how much the error depends on the output, which is the derivative of E in respect to $O_j$ (from (3)).

$$\frac{\partial E}{\partial O_j} = 2(O_j - d_j) \tag{6}$$

And then, how much the output depends on the activation, which in turn depends on the weights (from (1) and (2)):

$$\frac{\partial O_j}{\partial w_{ji}} = \frac{\partial O_j}{\partial A_j} \frac{\partial A_j}{\partial w_{ji}} = O_j(1 - O_j)x_i \tag{7}$$

And we can see that (from (6) and (7)):

$$\frac{\partial E}{\partial w_{ji}} = \frac{\partial E}{\partial O_j} \frac{\partial O_j}{\partial w_{ji}} = 2(O_j - d_j)O_j(1 - O_j)x_i \tag{8}$$

And so, the adjustment to each weight will be (from (5) and (8)):

$$\Delta w_{ji} = -2\eta(O_j - d_j)O_j(1 - O_j)x_i \tag{9}$$

We can use (9) as it is for training an ANN with two layers. Now, for training the network with one more layer we need to make some considerations. If we want to adjust the weights (let's call them $v_{ik}$) of a previous layer, we need first to calculate how the error depends not on the weight, but in the input from the previous layer. This is easy, we would just need to change $x_i$ with $w_{ji}$ in (7), (8), and (9). But we also need to see how the error of the network depends on the adjustment of $v_{ik}$. So:

$$\Delta v_{ik} = -\eta \frac{\partial E}{\partial v_{ik}} = -\eta \frac{\partial E}{\partial x_i} \frac{\partial x_i}{\partial v_{ik}} \qquad (10)$$

Where:

$$\frac{\partial E}{\partial w_{ji}} = 2(O_j - d_j)O_j(1 - O_j)w_{ji} \qquad (11)$$

And, assuming that there are inputs $u_k$ into the neuron with $v_{ik}$ (from (7)):

$$\frac{\partial x_i}{\partial v_{ik}} = x_i(1 - x_i)v_{ik} \qquad (12)$$

If we want to add yet another layer, we can do the same, calculating how the error depends on the inputs and weights of the first layer. We should just be careful with the indexes, since each layer can have a different number of neurons, and we should not confuse them.

For practical reasons, ANNs implementing the backpropagation algorithm do not have too many layers, since the time for training the networks grows exponentially. Also, there are refinements to the backpropagation algorithm which allow a faster learning.

**4.1. Exercise**

If you know how to program, implement the backpropagation algorithm, that at least will train the following network. If you can do a general implementation of the backpropagation algorithm, go ahead (for any number of neurons per layer, training sets, and even layers).

If you do not know how to program, but know how to use a mathematical assistant (such as Matlab or Mathematica), find weights which will suit the following network after defining functions which will ease your task.

If you do not have any computing experience, find the weights by hand.

The network for this exercise has three neurons in the input layer, two neurons in a hidden layer, and three neurons in the output layer. Usually networks are trained with large training sets, but for this exercise, we will only use one training example. When the inputs are (1, 0.25, -0.5), the outputs should be (1,-1,0). Remember you start with random weights.

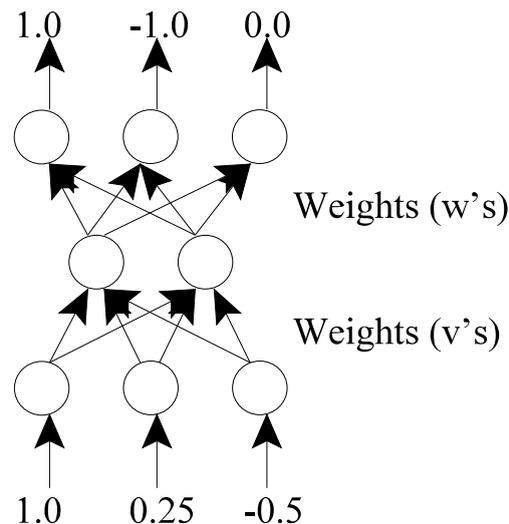

## 5. Further reading

The following great books go much deeper into ANNs:
* Rojas, R. (1996). *Neural Networks: A Systematic Introduction.* Springer, Berlin.
* Rumelhart, D. and J. McClelland (1986). *Parallel Distributed Processing*. MIT Press, Cambridge, Mass.

For further information on networks in general, and related themes, these books are quite useful and illustrative:
* Bar-Yam, Y. (1997). *Dynamics of Complex Systems*. Addison-Wesley.
* Kauffman, S. (1993) *Origins of Order*, Oxford University Press.

## 6. Online resources

There is a vast amount of resources on the Internet related to Neural Networks. A great tutorial, with excellent illustrative examples using Java applets (source code available), was developed at the EPFL (http://diwww.epfl.ch/mantra/tutorial/english/). Other two good tutorials are at the Universidad Politécnica de Madrid (http://www.gc.ssr.upm.es/inves/neural/ann1/anntutorial.html) and at the Imperial College of Science, Technology and Medicine University of London (http://www.doc.ic.ac.uk/~nd/surprise_96/journal/vol4/cs11/report.html). The author also has a small amount of resources related to programming neural networks in Java (http://jlagunez.iquimica.unam.mx/~carlos/programacione.html).